%% file: main.tex
\documentclass[runningheads]{llncs}
\usepackage[T1]{fontenc}
\usepackage{graphicx}
\usepackage{amsmath}
\usepackage{amssymb}
\usepackage{array}
\usepackage{wrapfig}
\usepackage{float}
\usepackage{placeins}
\begin{document}
\title{Guaranteed Conditional Diffusion: 3D Block-based Models for Scientific Data Compression}
\titlerunning{Guaranteed Conditional Diffusion Models}
%
\author{Jaemoon Lee \and
Xiao Li \and
Liangji Zhu \and
Sanjay Ranka 
\and Anand Rangarajan}
\authorrunning{J. Lee et al.}

%
\institute{University of Florida, Gainesville FL 32611, USA}
%
\maketitle              
\begin{abstract}
This paper proposes a new compression paradigm---Guaranteed Conditional Diffusion with Tensor Correction (GCDTC)---for lossy scientific data compression. The framework is based on recent conditional diffusion (CD) generative models, and it consists of a conditional diffusion model, tensor correction, and error guarantee. Our diffusion model is a mixture of 3D conditioning and 2D denoising U-Net. The approach leverages a 3D block-based compressing module to address spatiotemporal correlations in structured scientific data. Then, the reverse diffusion process for 2D spatial data is conditioned on the ``slices'' of content latent variables produced by the compressing module. After training, the denoising decoder reconstructs the data with zero noise and content latent variables, and thus it is entirely deterministic.
The reconstructed outputs of the CD model are further post-processed by our tensor correction and error guarantee steps to control and ensure a maximum error distortion, which is an inevitable requirement in lossy scientific data compression. 
Our experiments involving two datasets generated by climate and chemical combustion simulations show that our framework outperforms standard convolutional autoencoders and yields competitive compression quality with an existing scientific data compression algorithm.

\end{abstract}

\input{Narrative/01Introduction}

\input{Narrative/02RelatedWork}
\input{Narrative/03Methodology}
\input{Narrative/04Experiments}
\input{Narrative/05Conclusion}

\begin{credits}
\subsubsection{\ackname} This work was partially supported by DOE RAPIDS2 DE-SC0021320 and DOE DE-SC0022265.

\subsubsection{\discintname}
The authors have no competing interests to declare that are relevant to the content of this article.
\end{credits}

\input{Narrative/06Appendix}

%
%
%
\bibliographystyle{splncs04}
\bibliography{ref}

\end{document}

%% file: Narrative/01Introduction.tex
\section{Introduction}

Lossy scientific data compression has emerged as a vitally important area in the past decade. The volume and velocity of scientific data heighten the urgency of the requirement of good data compression algorithms, specifically methods that can provide performance guarantees in terms of error bounds on the primary data (PD) of interest. The concomitant rise of machine learning has seen the flowering of different learning-based compression paradigms. The primary ones are super-resolution, transform-based, and more recently methods based on generative AI. We first examine these paradigms before turning to the relatively new approaches based on generative models---the paradigm adopted in the present work.

Lossy compression based on \textbf{super-resolution} \cite{Khani2021,Conde2022swin2sr} is based on the premise that the data of interest can be faithfully reconstructed from a small set of ``true'' samples. Machine learning methods based on this paradigm attempt data reconstruction (of the original tensor) from this sample set. \textbf{Transform-based} methods have traditionally been the most popular paradigm with discrete cosine transforms (DCT), wavelets, principal component analysis (PCA) and dictionary-based methods leading the way. More recently, autoencoders (AE) which transform the data into a compact and quantized latent space from which learned decoders reconstruct the original tensor have been the paradigm of choice among ML practitioners. However, these paradigms do not leverage recent advances in generative AI. In this newer approach---termed \textbf{conditional diffusion (CD)} \cite{Yang2023cd}---the original tensor is first gradually converted into zero mean, Gaussian noise. Then, a decoder is learned which gradually denoises the tensor through stages to finally produce a tensor approximately drawn from the probability distribution of the original images. A latent space embedding is used to guide the diffusion process. We propose to work within this paradigm but in the context of scientific data compression.


\begin{wrapfigure}{r}{0.5\textwidth} 
\vspace{-0.4cm}
    \centering
    \includegraphics[width=\linewidth]{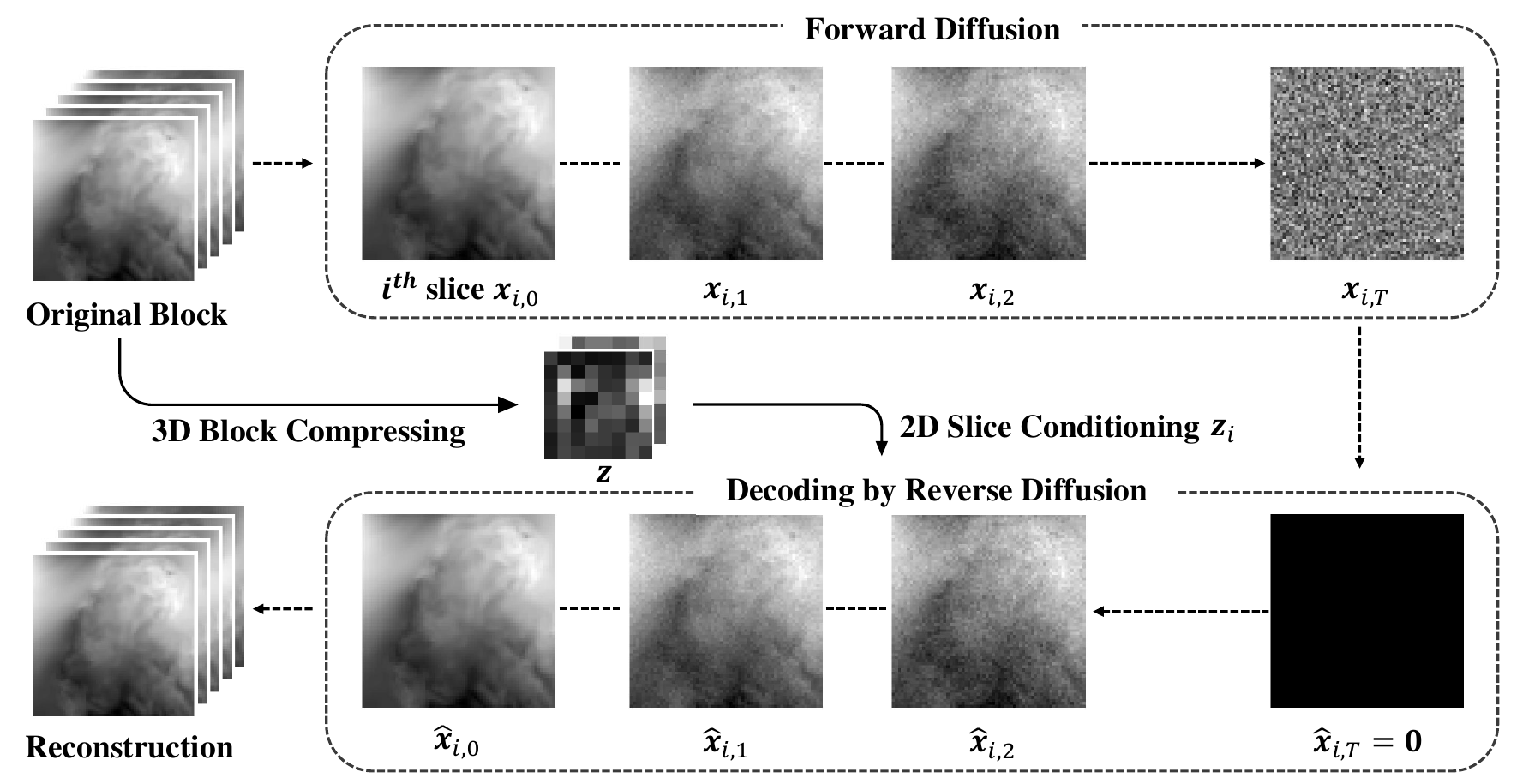}
  \caption{Overview of our conditional diffusion model for compression. We compress 3D blocks to capture spatiotemporal correlations in scientific datasets. The latent variables guide a 2D denoising diffusion process. Our denoising decoder reconstructs each of the 2D slices in 3D blocks based on its corresponding latent data $\boldsymbol{z}_i$. This enables us to keep a relatively simple U-Net architecture while getting effective conditioning via 3D block compression.}
    \label{fig:overview}
    \vspace{-0.4cm}
\end{wrapfigure}
 
Our approach to scientific data compression, situated within the conditional diffusion paradigm is now described. Figure~\ref{fig:overview} illustrates the overview of our proposed conditional diffusion models. Our model is a mixture of 3D block conditioning and 2D denoising diffusion. We first divide the original data into blocks of 3D tensors. 3D tensors are encoded into latent variables, which results in compressed codecs. We construct 3D latent embeddings using these codecs and they act as the conditioning information in CD. Unlike latent embeddings, we learn the denoising decoder in 2D space. Each 2D slice $\boldsymbol{x}_i$ of the 3D tensors is gradually converted into white noise in a stovepiped manner as described above. The denoising decoder estimates the noise of the 2D slice at the diffusion stage $t$. Using an embedding for the diffusion stage index, we learn the denoising decoder and the latent space embeddings in an end-to-end fashion. After training the machine, the original 3D tensor blocks are reconstructed with zero noise at the input (so that the decoder is entirely deterministic). The reconstructed primary data are examined to see if error bounds are violated and if so, we correct for the PD to be within pre-specified error bounds (using PCA or via a separate error bounding neural network). The main contributions are:
\begin{itemize}
    \item We propose a CD model for lossy scientific data compression. We divide the entire data into 3D tensors and map them into compressed codecs to capture spatiotemporal correlations in scientific datasets.
    \item The proposed CD model is a mixture of 3D conditioning and 2D diffusion. We prevent a complexity increase of our denoising decoder by avoiding a 3D diffusion process.
    \item To the best of our knowledge, this application of CD to scientific data compression with error guarantees---termed guaranteed conditional diffusion (GCD)---is a new contribution within a relatively new data compression paradigm.
\end{itemize}

%% file: Narrative/02RelatedWork.tex
\section{Related Work}

\subsubsection{Diffusion Models for Image Compression.}

Diffusion models \cite{Sohl2015,Song2019,Ho2020,Song2021scorebased,Nichol2021,Dhariwal2021,yu2024pet} have emerged as a dominant class of generative models in machine learning, demonstrating remarkable success across various domains, including image generation and natural language processing. Diffusion models also have recently been explored for image compression tasks. These works demonstrate how diffusion-based generative models can be adapted to achieve state-of-the-art performance in lossy image compression while maintaining high reconstruction quality.

Hoogeboom et al. proposed an autoregressive diffusion model (ADM) \cite{Hoogeboom2022autoregressive} to generate arbitrary-order data with order-agnostic autoregressive models and discrete diffusion models. They showed that ADM could be used for lossless compression tasks. Yang et al. \cite{Yang2023cd} proposed a lossy image compression framework using conditional diffusion models. Their approach transformed an image $\boldsymbol{x}_0$ into latent variables and produced embeddings used to guide a diffusion process of $\boldsymbol{x}_0$. Li et al. \cite{Li2024extreme} introduced a two-stage compression framework combining VAE-based encoding with pre-trained diffusion models for image reconstruction. The VAE-based module produced latent features that guided diffusion in latent space. The denoised latent features were processed by a decoder for reconstruction. Careil et al. \cite{Careil2024towards} employed iterative diffusion models for ultra-low bitrate image compression. Their two-branch architecture balances global structure and local texture, achieving state-of-the-art results in perceptual metrics like FID and KID. Relic et al. \cite{Relic2024lossy} reformulated quantization error removal as a denoising task, leveraging diffusion for latent recovery with minimal computation. Their codec outperforms traditional methods in realism metrics, maintaining high user preference even at reduced bitrates. Ma et al. \cite{ma2024correcting} developed a privileged end-to-end decoder using diffusion models, combining theoretical insights with convolutional decoders to enhance score function approximation. Their method achieves superior perceptual and distortion metrics compared to prior approaches.

These studies demonstrate the adaptability and promise of diffusion models for advancing image compression, setting the stage for future innovations. However, there is no work that leverages diffusion models in lossy scientific data compression.

%% file: Narrative/03Methodology.tex
\section{Methodology}

We propose a compression pipeline---Guaranteed Conditional Diffusion with Tensor Correction (GCDTC)---for lossy scientific data compression. The framework consists of three steps: (i) conditional diffusion, (ii) tensor correction, and (iii) error guarantee. Our approach leverages conditional generation of diffusion models \cite{Yang2023cd}, where compressed images guide the reverse diffusion process. In our framework, we introduce 3D block conditioning with 2D diffusion models. An overview of our conditional diffusion model is described in Figure~\ref{fig:overview}. Since scientific data have underlying temporal and spatial correlations, they must be addressed for effective data reduction. We first divide the entire dataset into 3D blocks and compress them to get codecs. Then, we use 2D slices of the compressed blocks for conditional generation. There is no advantage in using 3D diffusion models, as spatiotemporal correlations are already addressed by 3D block compression. After the conditional diffusion (CD) model is trained, the reconstructed output of the CD model is further enhanced by a tensor correction network followed by a error guaranteeing process. In this section, we briefly discuss the background of diffusion models and then describe our framework.

\subsection{Background}
\subsubsection{Diffusion models} are a class of hierarchical latent variable models that generate data by iteratively reversing a sequence of noise-adding transformations \cite{Sohl2015,Song2019,Ho2020,Song2021,yu2025robust}. These models define a joint distribution over data $\boldsymbol{x}_0$ and latent variables $\boldsymbol{x}_{1:T}$, where the latter represents intermediate noisy versions of the data. The generative process starts from a latent representation drawn from a simple prior (e.g., Gaussian noise) and progressively refines it to produce structured data, such as images.

The forward process $q$ systematically adds noise to the data through a series of Markovian transitions, eroding structure until the data becomes nearly indistinguishable from pure noise:
\begin{equation}\label{eq:forward}
    q\left(\boldsymbol{x}_t|\boldsymbol{x}_{t-1}\right)=\mathcal{N}\left(\boldsymbol{x}_t|\sqrt{1-\beta_t}\boldsymbol{x}_{t-1},\beta_t\boldsymbol{\mathrm{I}}\right),
\end{equation}
where $\beta_t$ is a predefined or learned noise schedule. The reverse process $p_\theta$ uses a neural network to iteratively recover the data by predicting the denoised states:
\begin{equation}\label{eq:reverse}
    p_\theta\left(\boldsymbol{x}_{t-1}|\boldsymbol{x}_t\right)=\mathcal{N}\left(\boldsymbol{x}_{t-1}|M_\theta\left(\boldsymbol{x}_t,t\right),\beta_t\boldsymbol{\mathrm{I}}\right).
\end{equation}

The reverse process is trained to approximate the true denoising transitions by minimizing a noise-prediction objective. Specifically, the model predicts the added noise $\epsilon$ for a given noisy sample $\boldsymbol{x}_t$, and the loss function is defined as:
\begin{equation}\label{eq:ddpm_loss}
    L\left(\theta,\boldsymbol{x}_0\right)=\mathbb{E}\left\|\epsilon-\epsilon_\theta\left(\boldsymbol{x}_t, t\right)\right\|^2,
\end{equation}
where $\boldsymbol{x}_t$ is constructed as $\boldsymbol{x}_t=\sqrt{\Bar{\alpha}_t}\boldsymbol{x}_0+\sqrt{1-\Bar{\alpha}_t}\epsilon$, $\alpha_t=1-\beta_t$, and $\Bar{\alpha}_t=\prod_{s=1}^t\left(\alpha_s\right)$. By leveraging this objective, the model learns to reverse the forward process and restore data from noise effectively.

For sampling, the reverse process can follow a stochastic trajectory, as in Denoising Diffusion Probabilistic Models (DDPMs) \cite{Ho2020}, or a deterministic path, as in Denoising Diffusion Implicit Models (DDIMs) \cite{Song2021}. The deterministic approach enables faster generation while retaining high-quality results, making it practical for tasks where efficiency is critical.

\subsubsection{Conditional diffusion models for compression}
extend diffusion models to lossy data compression by leveraging conditional generation \cite{Yang2023cd}. In this framework, an image $\boldsymbol{x}_0$ is compressed into a set of latent representations $\boldsymbol{z}$ using an entropy-optimized quantization process. These latent variables are then used as conditioning inputs to guide the reverse diffusion process:
\begin{equation}\label{eq:cd_reverse}
    p_\theta\left(\boldsymbol{x}_{t-1}|\boldsymbol{x}_t, \boldsymbol{z}\right)=\mathcal{N}\left(\boldsymbol{x}_{t-1}|M_\theta\left(\boldsymbol{x}_t, \boldsymbol{z}, t\right),\beta_t\boldsymbol{\mathrm{I}}\right).
\end{equation}
Here, $\boldsymbol{z}$ captures content information about an image, while the reverse diffusion process reconstructs finer texture details progressively. This approach combines the advantages of learned image compression approaches \cite{Balle2017,Balle2018,Minnen2018} with the generative capabilities of diffusion models to propose a new paradigm in lossy image compression. The model is trained using a similar objective to the DDPM loss function in~\eqref{eq:ddpm_loss}, but with an added conditioning mechanism for $\boldsymbol{z}$,
\begin{equation}\label{eq:cd_loss}
    L\left(\boldsymbol{x}_0|\boldsymbol{z}\right)=\mathbb{E}\left\|\epsilon-\epsilon_\theta\left(\boldsymbol{x}_t,\boldsymbol{z}, t\right)\right\|^2=\frac{\Bar{\alpha}_t}{1-\Bar{\alpha}_t}\mathbb{E}\left\|\boldsymbol{x}_0-\mathcal{X}_\theta\left(\boldsymbol{x}_t,\boldsymbol{z}, t\right)\right\|^2,
\end{equation}
where $\mathcal{X}$-prediction, denoted $\mathcal{X}_\theta$, is an equivalent alternative of the noise-based loss function and directly learns to reconstruct $\boldsymbol{x}_0$ instead of the noise $\epsilon$. At test time, the latent variable $\boldsymbol{z}$ is entropy-decoded, and denoising U-Net conditionally reconstructs the image $\boldsymbol{x}_0$ with the iterative reverse process. The sampling process in \cite{Yang2023cd} follows the DDIM's sampling method \cite{Song2021} with either a deterministic way with zero noise or a stochastic noise drawn from a normal distribution. This conditional setup allows the model to work suitably for compression, not generation.

\subsection{3D Conditional Diffusion Model for Scientific Data Compression}

\begin{figure}[ht]
    \centering
    \includegraphics[width=\textwidth]{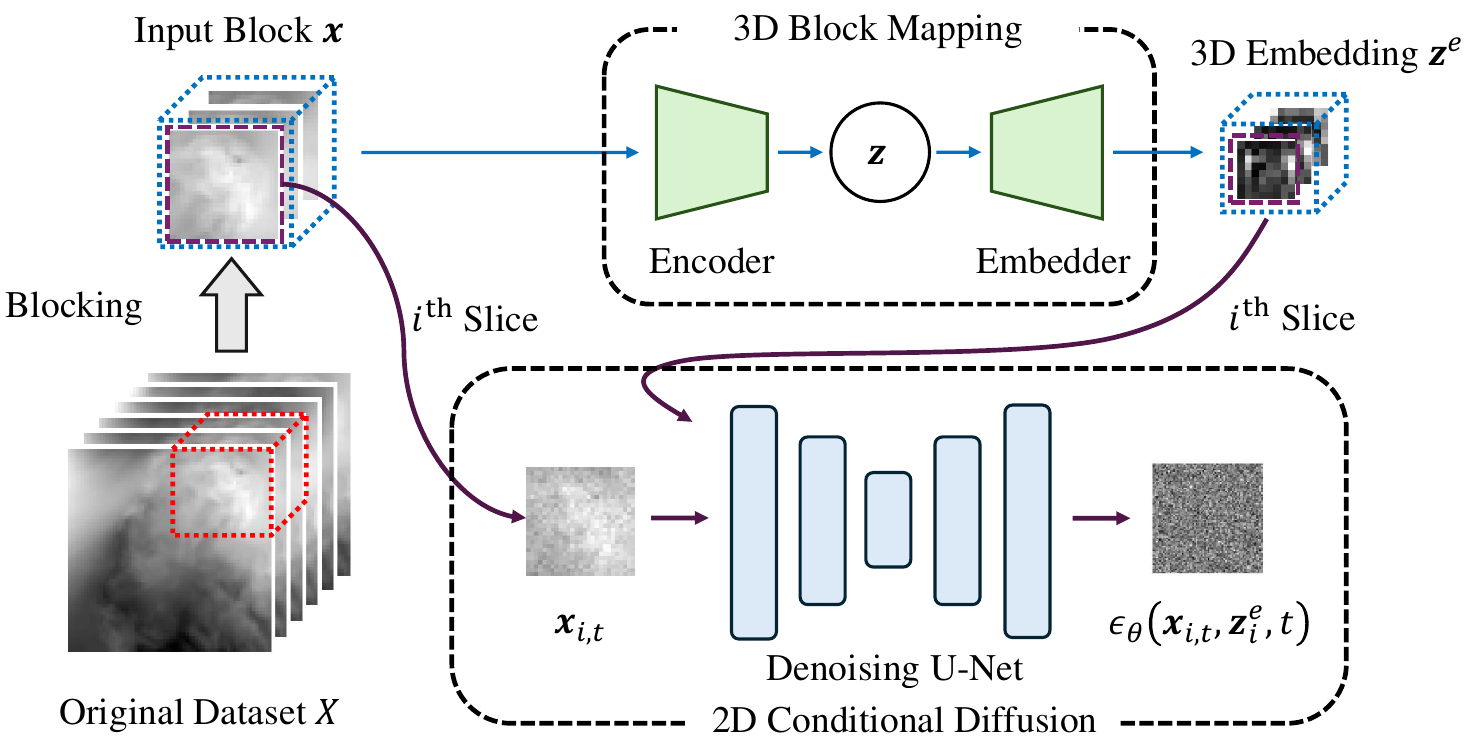}
    \caption{A visualization of our 3D conditional diffusion model. We obtain 3D embedding $\boldsymbol{z}^e$ from the tensor block $\boldsymbol{x}$ to effectively address spatiotemporal correlations in scientific datasets. If the tensor block size is $D\times H\times W$, the first dimension of $\boldsymbol{z}^e$ must be equal to $D$. This is because we incorporate 2D diffusion and the $i^\mathrm{th}$ slice $\boldsymbol{z}_i^e$ is used to condition the denoising decoder. Hence, the decoder learns to predict the 2D noise at the diffusion stage $t$ of each 2D slice $\boldsymbol{x}_i$ in $\boldsymbol{x}$. Architecture details are described in Appendix.}\label{fig:cd_model}
    \vspace{-0.4cm}
\end{figure}

Our approach is developed upon the conditional diffusion model \cite{Yang2023cd}, where latent information $\boldsymbol{z}$ guides the reverse diffusion process for 2D image compression. We extend the conditional diffusion model to 3D scientific data compression, which is a mixture of a 3D block encoder and a conditional 2D denoising U-Net as described in Figure~\ref{fig:cd_model}.
\vspace{-0.4cm}
\subsubsection{Training.}
We first divide the entire dataset into a set of 3D tensors. The tensors are mapped into latent variable $\boldsymbol{z}$ and converted into 3D embedding $\boldsymbol{z}^e$. In the diffusion process, we follow the same method of the DDPM in \eqref{eq:forward} to get the 2D noisy slice $\boldsymbol{x}_{i,t}$ at the stage $t$. Using the slice of the latent embedding $\boldsymbol{z}_i^e$, we modify the conditional reverse diffusion process in \eqref{eq:cd_reverse} as
\begin{equation}\label{eq:our_reverse}
    p_\theta\left(\boldsymbol{x}_{i,t-1}|\boldsymbol{x}_{i,t}, \boldsymbol{z}_i^e\right)=\mathcal{N}\left(\boldsymbol{x}_{i,t-1}|M_\theta\left(\boldsymbol{x}_{i,t}, \boldsymbol{z}_i^e, t\right),\beta_t\boldsymbol{\mathrm{I}}\right).
\end{equation}
Like in \eqref{eq:cd_loss}, our model is trained to estimate the noise,
\begin{equation}\label{eq:our_loss}
    L\left(\boldsymbol{x}_{i,0}|\boldsymbol{z}_i^e\right)=\mathbb{E}\left\|\epsilon-\epsilon_\theta\left(\boldsymbol{x}_{i,t},\boldsymbol{z}_i^e, t\right)\right\|^2.
\end{equation}
With the diffusion stage embeddings, the 2D denoising decoder and 3D latent embeddings are trained in an end-to-end manner.

\subsubsection{Quantization and Decoding.}
After the model is trained, we quantize and entropy-encode $\boldsymbol{z}$ using scalar quantization followed by Huffman encoding. Each value $z$ in $\boldsymbol{z}$ is rounded by $\left\lfloor \frac{b}{a}z\right\rfloor$, where $b$ is a power of 10 and $a$ determines the quantization bin size. Then, all the integers are Huffman-encoded to produce compressed codecs. In learned image compression \cite{Balle2017,Balle2018,Minnen2018}, a prior $p\left(\boldsymbol{z}\right)$ is modeled as an additional neural network to include quantization and entropy coding for end-to-end learning. We don't incorporate hyper-prior networks because of (1) additional steps in our framework that complement the output of the CD model and (2) reducing the model complexity.

At decoding time, we entropy-decode and dequantize the compressed codecs and produce latent embeddings. The denoising decoder reconstructs the data deterministically with $\boldsymbol{x}_{i,T}=\boldsymbol{0}$. We reconstruct a 3D tensor $\hat{\boldsymbol{x}}_0$ using ancestral sampling for each 2D slice $\boldsymbol{x}_{i}$ in $\boldsymbol{x}_0$. We follow the DDPM's sampling method,
\begin{equation}
    \boldsymbol{x}_{i,t-1}=\frac{1}{\sqrt{\alpha_t}}\left(\boldsymbol{x}_{i,t}-\frac{1-\alpha_t}{\sqrt{1-\Bar{\alpha}_t}}\epsilon_\theta\left(\boldsymbol{x}_{i,t},\boldsymbol{z}_i^e, t\right)\right).
\end{equation}

\subsection{Tensor Correction and Error Guarantee}\label{subsec:tc_eg}
In lossy scientific data compression, the maximum distortion of reconstructed raw data or primary data (PD) is required to be bounded by a user-specific error target due to reliable post-analysis. For this reason, error-bounded lossy compression is dominant in this field. These techniques are developed upon several methods based on prediction (SZ \cite{SZ3}), transformation (ZFP \cite{ZFP2}), and wavelet decomposition (MGARD \cite{MGARD_3} and SPERR \cite{SPERR}). To provide guaranteed reconstruction, we first enhance the output of the CD model and then correct tensors.
\vspace{-0.6cm}
\subsubsection{Tensor Correction.}
We incorporate a tensor correction (TC) network proposed in \cite{JL-S3D_arxiv}. The TC network is trained using reconstructed tensors obtained from the CD model, and it is an ``overcomplete'' feedforward network to learn a reverse mapping from the reconstructed tensor $\hat{\boldsymbol{x}}$ to the original tensor $\boldsymbol{x}$. In our TC network, we use tensors along the perpendicular direction to 2D slices in the CD model. The TC network reproduces the enhanced reconstructed dataset $\hat{X}$.
\vspace{-0.2cm}
\subsubsection{Error Guarantee.}
We utilize a post-processing method to control a maximum error distortion according to a pre-specified error bound \cite{JL-GAE,JL-S3D_arxiv}. We first divide the original and reconstructed datasets into small blocks ($\{\boldsymbol{b}\}_{i=1}^N$ and $\{\hat{\boldsymbol{b}}\}_{i=1}^N$) to bound errors on this small region. We perform PCA on the residual set $\{\boldsymbol{b}-\hat{\boldsymbol{b}}\}_{i=1}^N$ to get a global basis $U$. The error guarantee step is a selective process. For each block $\boldsymbol{b}$ that exceeds the error bound $\tau$ ($\|\boldsymbol{b}-\hat{\boldsymbol{b}}\|_{2}> \tau$), we draw a coefficient variable $\boldsymbol{c}$,
\begin{equation}
    \boldsymbol{c}=U^T\left(\boldsymbol{b}-\hat{\boldsymbol{b}}\right).
\end{equation}
The coefficient $\boldsymbol{c}$ is quantized using log-based scalar quantization \cite{JL-S3D_arxiv}. We dequantize the coefficient and sort its entries according to their contribution to the $\ell_2$ error of $\hat{\boldsymbol{b}}$. We select the optimal subset $\boldsymbol{c}_s$ in $\boldsymbol{c}$ and its corresponding basis set $U_s$ to satisfy the target error bound. The final corrected block $\Tilde{\boldsymbol{b}}$ is
\begin{equation}
    \Tilde{\boldsymbol{b}}=\hat{\boldsymbol{b}}+U_{s}\boldsymbol{c}_{s}.
\end{equation}
The selected coefficients and basis indices are Huffman-encoded. In our framework, the compressed codecs include all the encoded data of latent variables in the CD model and selected coefficients with their indices. We also consider all the sizes of models, PCA basis, and dictionaries as compression costs. Otherwise, we can use an extremely large machine that is even larger than the given scientific dataset, and can do overfitting to produce ideal results. This is because scientific datasets are domain-specific, unlike natural image compression.

%% file: Narrative/04Experiments.tex
\section{Experiments}
This section presents the experimental results of our compression framework---Guaranteed Conditional Diffusion with Tensor Correction (GCDTC). We utilize two scientific datasets generated by E3SM and S3D applications. All the experiments are conducted using an Nvidia K80 GPU in OLCF's Andes.

\subsection{Datasets, Metrics, and Baselines}

\subsubsection{E3SM Dataset.}
We use the dataset generated by the E3SM (Energy Exascale Earth System Model) \cite{e3sm} that simulates Earth's climate system. For each 30-day period, the E3SM simulates climate variables of $240\times 240$ resolution with float32 data points in 6 regions with 720 timesteps. We use the cropped sea-level pressure (PSL) climate variable for 3 months, which results in a dataset with dimensions of $6\times 2160\times 192 \times 192$. We construct 3D blocks across temporal and spatial spaces in each region.
\vspace{-0.2cm}
\subsubsection{S3D Dataset.}
The dataset is generated by Sandia’s compressible reacting direct numerical simulation (DNS) code, S3D \cite{s3d}. The S3D simulates chemically reacting flow, involving detailed chemical mechanisms across numerous species. We use three species' $640\times 640$ mass fraction with float64 data points, collected over 288 timesteps, and create 3D blocks for each species.

\vspace{-0.2cm}
\subsubsection{Metrics.}
We use the \textit{NRMSE} and \textit{compression ratio} for the compression quality evaluation. NRMSE is the normalized RMSE, where RMSE is divided by a data range. In the computation of compression ratios, we consider all the sizes of models and dictionaries for entropy coding as scientific compression techniques are applied to a specific scientific application. Otherwise, we might use an extremely large machine, even bigger than a dataset, which can produce ideal compression results using overfitting.

\vspace{-0.2cm}
\subsubsection{Baselines.}
We compare our method to SZ3 \cite{SZ3} and a standard convolutional autoencoder (AE). SZ is one of the most dominant error-bounded lossy compressors for lossy scientific data compression. It is a prediction-based method, where a data point is estimated by its neighbors. The prediction accuracy is affected by a specified point-wise error bound. For the autoencoder, we incorporate 3D convolutional layers. The architecture is almost the same as the encoder and embedder structures of the conditional diffusion model, described in the Appendix. The only difference is the number of the output channels in the last unit. The model is trained to minimize the MSE loss between the original and reconstructed data. We then apply the error guarantee process in Section~\ref{subsec:tc_eg} to the output of the autoencoder.

\subsection{Implementation Details}
\subsubsection{Training.}
We don't split the datasets into training and test sets as error-bounded lossy compressors including SZ are not learned models. We divide each dataset into a set of $16\times 64\times 64$ blocks. The model is trained using the Adam optimizer \cite{adam} with the learning rate of $1\times 10^{-3}$ and 100 epochs. The number of diffusion steps is 1000 and the linear noise scheduling method of the DDPM is incorporated. The maximum and minimum $\beta$ are set to $5\times 10^{-3}$ and $1\times 10^{-5}$ respectively. Our framework is implemented using Pytorch \cite{pytorch}. The architecture detail of our conditional diffusion model is illustrated in the Appendix. We use the tensor correction network in \cite{JL-S3D_arxiv}. The inputs are 60-dimensional and 48-dimensional tensors across temporal space in E3SM and S3D respectively. In the error guarantee process in Section~\ref{subsec:tc_eg}, we correct $4\times 4\times 4$ blocks for both E3SM and S3D. The quantization factor $b$ and bin size $a$ are set to 1000 and 16 respectively for both latent variables $\boldsymbol{z}$ and selected coefficients $\boldsymbol{c}_s$.

\subsection{Results and Discussion}
\begin{figure}[H]  
    \centering
    \includegraphics[width=0.4\textwidth]{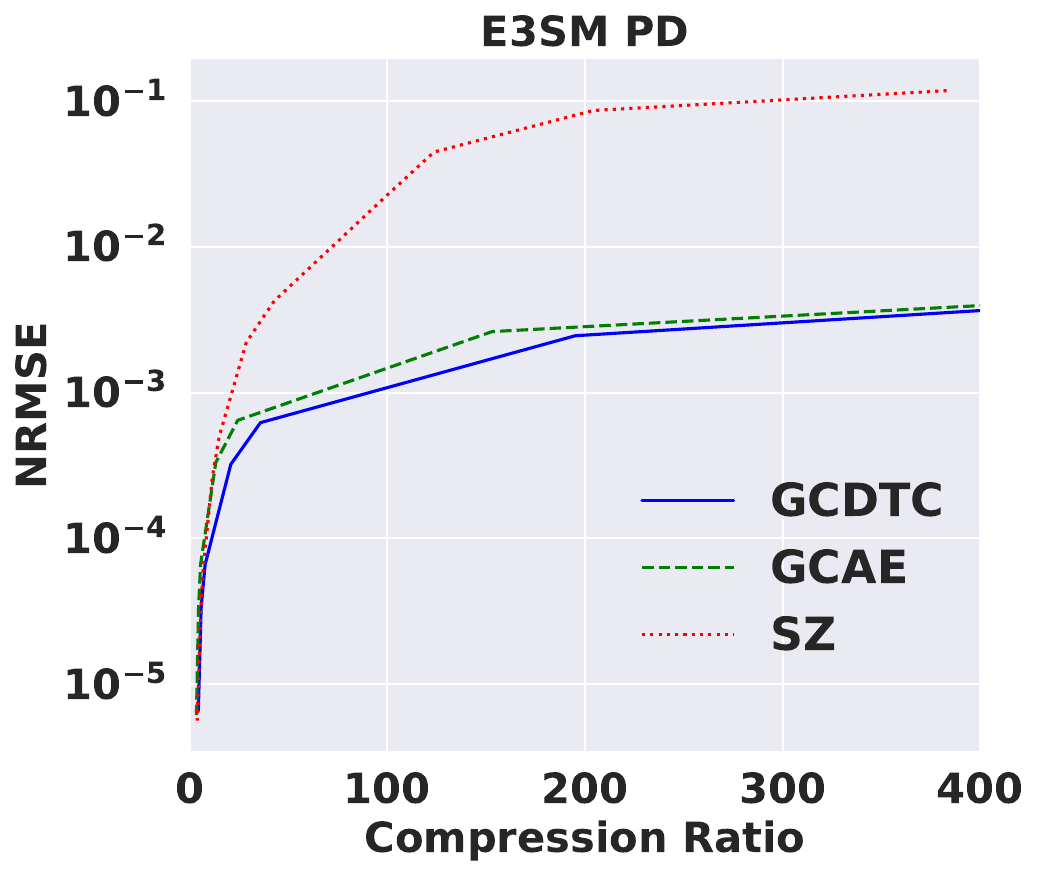}
    \hfill
    \includegraphics[width=0.4\textwidth]{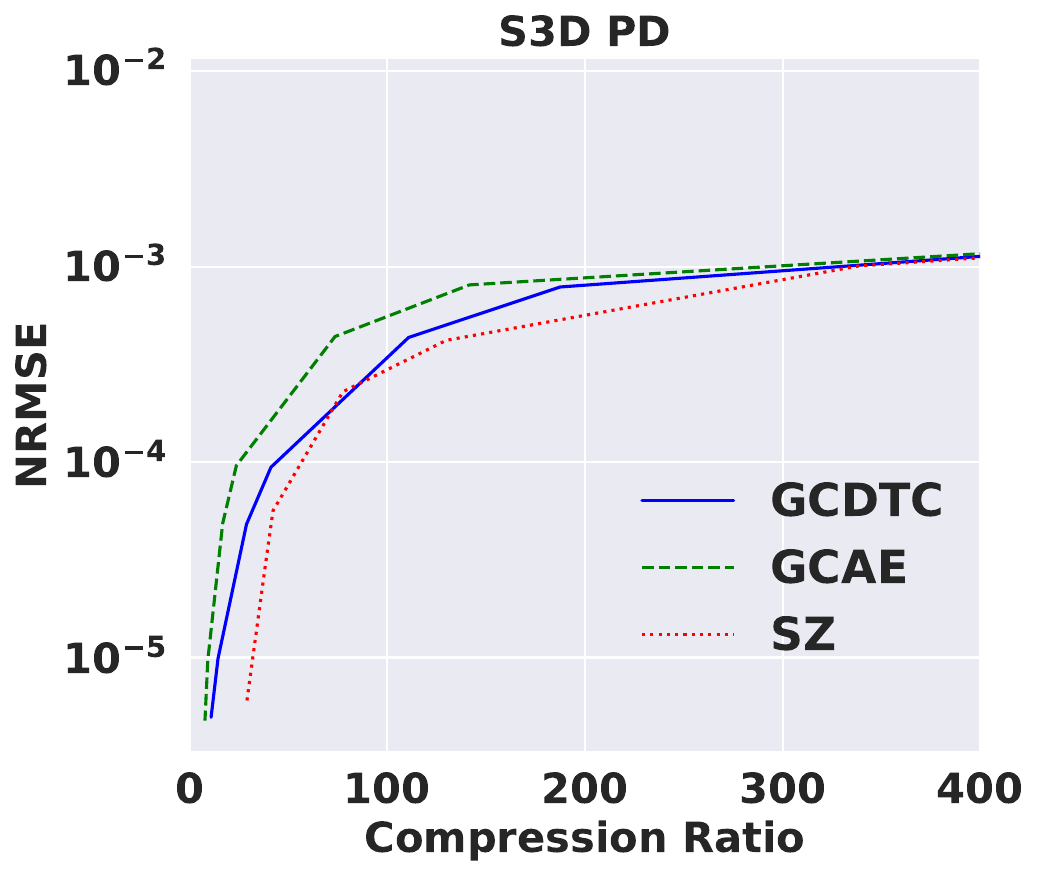}
    \caption{Reconstruction quality vs. compression ratio evaluation on E3SM (left) and S3D (right) datasets. GCDTC and GCAE denote Guaranteed Conditional Diffusion with Tensor Correction and Guaranteed Convolutional AutoEncoder. Note that the NRMSE results (y-axis) are plotted on a log scale. The result shows that our GCDTC outperforms GCAE, while yielding competitive performance with SZ.}
    \label{fig:compress_result}
\end{figure}

\begin{figure}[H]  
    \centering
    \includegraphics[width=\textwidth]{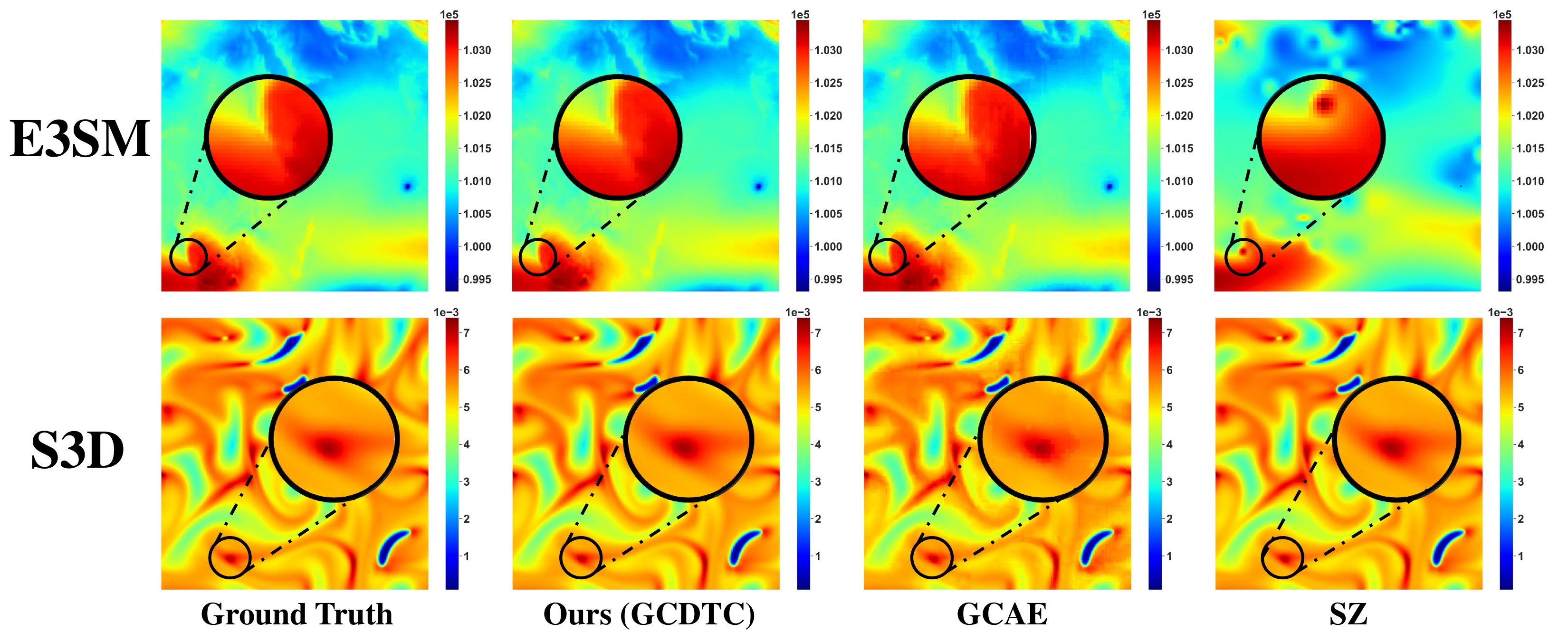}
    \caption{Visualization of reconstructions in E3SM and S3D at compression ratio 100.}
    \label{fig:compress_example}
\end{figure}

We compress and reconstruct the datasets, denoted as PD. We vary the error bounds, the maximum point-wise distortion (SZ) and region-wise distortion (Ours), to get various compression results as shown in Figure~\ref{fig:compress_result}. Our experiments reveal that our proposed GCDTC outperforms the standard convolutional autoencoder in both the E3SM and S3D datasets. In the conditional diffusion model, latent variables convey content information and the denoising decoder processes further details. Both of them contribute reconstructions, which results in better compression quality compared to the autoencoder. Compared to SZ, GCDTC achieves at least twice larger compression ratios above $1\times 10^{-4}$ NRMSE in E3SM, while yielding competitive compression results in S3D. In lossy scientific data compression, $1\times 10^{-3}$ NRMSE is usually used as a target or an acceptable compression quality level for post-analysis, and GCDTC shows decent amounts of data reduction at this NRMSE level. Figure~\ref{fig:compress_example} shows reconstruction examples at the compression ratio 100. We zoom into a small region to check the details. Our framework can capture the details of the original data.

\FloatBarrier  

\FloatBarrier  

We also evaluate the complexity by comparing the number of model parameters and decoding time. The machine sizes are included in the compression ratio computation. GCDTC suffers from slow decoding speed due to the iterative denoising process. We will address this limitation by incorporating progressive distillation \cite{Salimans2022progressive} that reduces the number of iterations in our future work. Despite this limitation, our conditional diffusion model proves its effectiveness and introduces the recent paradigm in generative AI in the context of lossy scientific data compression.

\begin{table}[H]  
\centering
\caption{Model complexity and decoding time. Acronyms are in Figure~\ref{fig:compress_result}.}
\label{tab:complexity}
\begin{tabular}{| >{\centering}m{3.5cm}| >{\centering}m{2.5cm}| >{\centering}m{2.5cm}|>{\centering\arraybackslash}m{2.5cm}|}
\hline
\textbf{Model} & \textbf{GCDTC (Ours)} & \textbf{GCAE} & \textbf{SZ} \\
\hline
Number of Parameters & 1.5 M & 1.0 M & - \\
\hline
Decoding Time (sec) & 978.2 & 2.2 & 12.2 \\
\hline
\end{tabular}
\end{table}

%% file: Narrative/05Conclusion.tex
\section{Conclusions}

In this work, we have shown the efficacy of a generative AI model---guaranteed conditional diffusion (GCD)---for scientific data compression. In contrast to traditional video, scientific data comprise blocks of tensors and therefore we designed GCD to work in this setting. GCD has three modules: (i) the standard 2D diffusion architecture with U-Net to learn each denoising stage, (ii) an encoder which compresses 3D blocks, produces latent variables, and acts as the conditioning information, and (iii) a post-processing tensor correction network which provides error bound guarantees at each compression ratio. Results on two applications---a climate dataset (E3SM) and a CFD dataset (S3D)---demonstrate that GCD is better or competitive with SZ---a standard lossy scientific data compression method and convolutional autoencoders. Future work will center on four aspects which are expected to further improve GCD: (i) a full adaptation to blocks and hyper-blocks of tensors, (ii) the use of a scale hyperprior approach for quantization, (iii) incorporation of attention within U-Net, and (iv) fast decoding via distillation. We eventually expect GCD and its variants to be a popular third paradigm for data compression, taking its place alongside super-resolution and transform-based paradigms.

%% file: Narrative/06Appendix.tex
\section*{\appendixname}

\begin{figure}[ht]
    \vspace{-0.4cm}
    \centering
    \includegraphics[width=\textwidth]{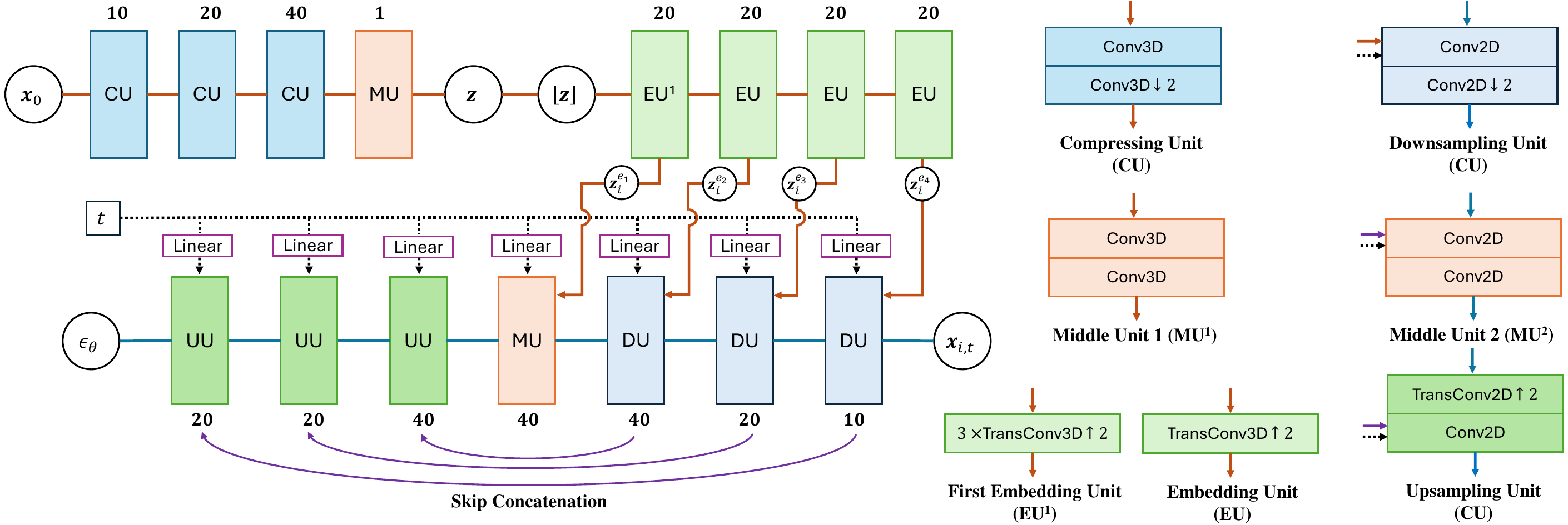}
    \caption{Illustration of our conditional diffusion model architecture. Numbers above or below the units indicate output channels.}\label{fig:cd_archit}
    \vspace{-0.8cm}
\end{figure}

\paragraph{\textbf{Architectures.}}
We follow a U-Net design in our denoising decoder with simple units to reduce the number of parameters. The key idea is to produce a 3D latent embedding $\boldsymbol{z}_i^e$ and use 2D slices for conditioning. To do so, the first embedding unit, denoted as $\textrm{EU}^1$ in Figure~\ref{fig:cd_archit}, constructs $\boldsymbol{z}_{i}^{e_1}$ such that the number of 2D slices in $\boldsymbol{z}_{i}^{e_1}$ becomes equal to that of  2D slices in $\boldsymbol{x}_0$. In this way, the 2D slices of latent embeddings can be used to condition the 2D diffusion process.

%% file: main.bbl
\begin{thebibliography}{10}
\providecommand{\url}[1]{\texttt{#1}}
\providecommand{\urlprefix}{URL }
\providecommand{\doi}[1]{https://doi.org/#1}

\bibitem{MGARD_3}
Ainsworth, M., Tugluk, O., Whitney, B., Klasky, S.: Multilevel techniques for compression and reduction of scientific data-quantitative control of accuracy in derived quantities. SIAM Journal on Scientific Computing  \textbf{41}(4),  A2146--A2171 (2019)

\bibitem{Balle2017}
Ball{\'e}, J., Laparra, V., Simoncelli, E.P.: {End-to-end Optimized Image Compression}. In: International Conference on Learning Representations (2017)

\bibitem{Balle2018}
Ballé, J., et~al.: Variational image compression with a scale hyperprior. In: International Conference on Learning Representations (2018)

\bibitem{Careil2024towards}
Careil, M., Muckley, M.J., Verbeek, J., Lathuili{\`e}re, S.: {Towards image compression with perfect realism at ultra-low bitrates}. In: The Twelfth International Conference on Learning Representations (2024)

\bibitem{Conde2022swin2sr}
Conde, M.V., Choi, U.J., Burchi, M., Timofte, R.: {Swin2sr: Swinv2 transformer for compressed image super-resolution and restoration}. In: European Conference on Computer Vision. pp. 669--687 (2022)

\bibitem{Dhariwal2021}
Dhariwal, P., Nichol, A.: {Diffusion Models Beat GANs on Image Synthesis}. In: Advances in Neural Information Processing Systems. vol.~34, pp. 8780--8794 (2021)

\bibitem{ZFP2}
Diffenderfer, J., Fox, A.L., Hittinger, J.A., Sanders, G., Lindstrom, P.G.: {Error Analysis of ZFP Compression for Floating-Point Data}. SIAM Journal on Scientific Computing  \textbf{41}(3),  A1867--A1898 (2019)

\bibitem{e3sm}
Golaz, J.C., Caldwell, P.M., et~al.: {The DOE E3SM Coupled Model Version 1: Overview and Evaluation at Standard Resolution}. Journal of Advances in Modeling Earth Systems  \textbf{11}(7),  2089--2129 (2019)

\bibitem{Ho2020}
Ho, J., Jain, A., Abbeel, P.: {Denoising Diffusion Probabilistic Models}. In: Advances in Neural Information Processing Systems. vol.~33, pp. 6840--6851 (2020)

\bibitem{Hoogeboom2022autoregressive}
Hoogeboom, E., et~al.: {Autoregressive Diffusion Models}. In: International Conference on Learning Representations (2022)

\bibitem{Khani2021}
Khani, M., Sivaraman, V., Alizadeh, M.: {Efficient Video Compression via Content-Adaptive Super-Resolution}. In: Proceedings of the IEEE/CVF International Conference on Computer Vision (ICCV). pp. 4521--4530 (2021)

\bibitem{adam}
Kingma, D.P., Ba, J.: {Adam: {A} Method for Stochastic Optimization}. In: 3rd International Conference on Learning Representations, {ICLR} 2015 (2015)

\bibitem{JL-S3D_arxiv}
Lee, J., Jung, K.S., Gong, Q., Li, X., Klasky, S., Chen, J., Rangarajan, A., Ranka, S.: {Machine Learning Techniques for Data Reduction of CFD Applications} (2024)

\bibitem{JL-GAE}
Lee, J., Rangarajan, A., Ranka, S.: {Nonlinear-by-Linear: Guaranteeing Error Bounds in Compressive Autoencoders}. In: Proceedings of the 2023 Fifteenth International Conference on Contemporary Computing. p. 552–561 (2023)

\bibitem{SPERR}
Li, S., Lindstrom, P., Clyne, J.: {Lossy Scientific Data Compression With SPERR}. In: 2023 IEEE International Parallel and Distributed Processing Symposium (IPDPS). pp. 1007--1017 (2023)

\bibitem{Li2024extreme}
Li, Z., Zhou, Y., Wei, H., Ge, C., Jiang, J.: {Towards Extreme Image Compression with Latent Feature Guidance and Diffusion Prior} (2024)

\bibitem{SZ3}
Liang, X., et~al.: {SZ3: A Modular Framework for Composing Prediction-Based Error-Bounded Lossy Compressors}. IEEE Transactions on Big Data  (2023)

\bibitem{ma2024correcting}
Ma, Y., Yang, W., Liu, J.: {Correcting Diffusion-Based Perceptual Image Compression with Privileged End-to-End Decoder}. In: Forty-first International Conference on Machine Learning (2024)

\bibitem{Minnen2018}
Minnen, D., Ball\'{e}, J., Toderici, G.D.: {Joint Autoregressive and Hierarchical Priors for Learned Image Compression}. In: Advances in Neural Information Processing Systems. vol.~31 (2018)

\bibitem{Nichol2021}
Nichol, A.Q., Dhariwal, P.: {Improved Denoising Diffusion Probabilistic Models} (2021)

\bibitem{pytorch}
Paszke, A., Gross, S., et~al.: {PyTorch: An Imperative Style, High-Performance Deep Learning Library}. In: Advances in Neural Information Processing Systems. vol.~32 (2019)

\bibitem{Relic2024lossy}
Relic, L., Azevedo, R., Gross, M., Schroers, C.: {Lossy Image Compression with Foundation Diffusion Models}. In: European Conference on Computer Vision (ECCV) (2024)

\bibitem{Salimans2022progressive}
Salimans, T., Ho, J.: {Progressive Distillation for Fast Sampling of Diffusion Models}. In: International Conference on Learning Representations (2022)

\bibitem{Sohl2015}
Sohl-Dickstein, J., Weiss, E., Maheswaranathan, N., Ganguli, S.: {Deep Unsupervised Learning using Nonequilibrium Thermodynamics}. In: Proceedings of the 32nd International Conference on Machine Learning. vol.~37, pp. 2256--2265 (2015)

\bibitem{Song2021}
Song, J., Meng, C., Ermon, S.: {Denoising Diffusion Implicit Models}. In: International Conference on Learning Representations (2021)

\bibitem{Song2019}
Song, Y., Ermon, S.: Generative modeling by estimating gradients of the data distribution. In: Proceedings of the 33rd International Conference on Neural Information Processing Systems (2019)

\bibitem{Song2021scorebased}
Song, Y., Sohl-Dickstein, J., Kingma, D.P., Kumar, A., Ermon, S., Poole, B.: {Score-Based Generative Modeling through Stochastic Differential Equations}. In: International Conference on Learning Representations (2021)

\bibitem{Yang2023cd}
Yang, R., Mandt, S.: {Lossy Image Compression with Conditional Diffusion Models}. In: Advances in Neural Information Processing Systems. vol.~36 (2023)

\bibitem{s3d}
Yoo, C.S., Lu, T., Chen, J.H., Law, C.K.: {Direct numerical simulations of ignition of a lean $n-$heptane/air mixture with temperature inhomogeneities at constant volume: Parametric study}. Combust. Flame  \textbf{158},  1727--1741 (2011)

\bibitem{yu2025robust}
Yu, B., Ozdemir, S., Dong, Y., Shao, W., Pan, T., Shi, K., Gong, K.: Robust whole-body pet image denoising using 3d diffusion models: evaluation across various scanners, tracers, and dose levels. European Journal of Nuclear Medicine and Molecular Imaging pp. 1--14 (2025)

\bibitem{yu2024pet}
Yu, B., Ozdemir, S., Dong, Y., Shao, W., Shi, K., Gong, K.: Pet image denoising based on 3d denoising diffusion probabilistic model: Evaluations on total-body datasets. In: International Conference on Medical Image Computing and Computer-Assisted Intervention. pp. 541--550. Springer (2024)

\end{thebibliography}
